\documentclass{article}

\usepackage{PRIMEarxiv}

\usepackage[utf8]{inputenc} 
\usepackage[T1]{fontenc} 
\usepackage{booktabs}       
\usepackage{amsfonts}       
\usepackage{nicefrac}       
\usepackage{microtype}      
\usepackage{xcolor}         
\usepackage{makecell}
\usepackage{url} 
\usepackage{microtype}
\usepackage{graphicx}
\usepackage{subfigure}
\usepackage{booktabs}
\usepackage{hyperref}
\hypersetup{colorlinks,linkcolor={blue},citecolor={blue},urlcolor={blue}}
\usepackage{wrapfig}
\usepackage{pythonhighlight}
\usepackage{xcolor}
\usepackage{booktabs}       
\usepackage{amsfonts}       
\usepackage{nicefrac}       
\usepackage{microtype}      
\usepackage{lipsum}
\usepackage{fancyhdr}       
\usepackage{graphics}
\usepackage{epsfig}
\usepackage{amsmath}
\usepackage{multirow}
\usepackage{threeparttable}
\usepackage{caption}
\usepackage{listings}
\usepackage{mathabx}
\usepackage{graphicx}       
\graphicspath{{media/}}     

\newcommand{\ours}{SMILE}

\title{SMILE: Scaling Mixture-of-Experts with Efficient Bi-level Routing 
}
\author{%
  Chaoyang He$^1$, Shuai Zheng$^2$, Aston Zhang$^2$, George Karypis$^2$, Trishul Chilimbi$^2$\\ \bf Mahdi Soltanolkotabi$^1$, Salman Avestimehr$^1$ \\
  $^1$University of Southern California\\
  $^2$AWS AI\\
  Email : 
\texttt{chaoyang.he@usc.edu} \\
}

\begin{document}
\maketitle

\begin{abstract}
The mixture of Expert (MoE) parallelism is a recent advancement that scales up the model size with constant computational cost. MoE selects different sets of parameters (i.e., experts) for each incoming token, resulting in a sparsely-activated model. Despite several successful applications of MoE, its training efficiency degrades significantly as the number of experts increases. The routing stage in MoE relies on the efficiency of the All2All communication collective, which suffers from network congestion and has poor scalability. To mitigate these issues, we introduce \ours, which exploits heterogeneous network bandwidth and splits a single-step routing into bi-level routing. Our experimental results show that the proposed method obtains a 2.5$\times$ speedup over Switch Transformer in terms of pretraining throughput on the “Colossal Clean Crawled Corpus” without losing any convergence speed.
\end{abstract}


\section{Introduction}
\label{sec:intro}

Gigantic models have recently gained significant attention due to their remarkable performance on natural language processing \cite{brown2020language}, computer vision \cite{dosovitskiy2020image}, and cross-modal learning \cite{radford2021learning}. However, the training of gigantic models requires significant computational resources and data. As the model size is scaled up, such large-scale training becomes computationally intensive and environmentally unfriendly \cite{patterson2021carbon}. In perspective, the total emissions of net tCO2e for GPT-3 are around 552 tCO2e \cite{patterson2021carbon}, while a direct round trip of a single passenger jet between San Francisco and New York emits about 180 tCO2e \cite{patterson2021carbon}. Recent studies have started seeking alternative approaches to enable greater computational efficiency.

Mixture of Experts (MoEs) \cite{shazeer2017outrageously} have emerged as the foundational neural network to scale up model capacity using a massive number of parameters while maintaining a constant computational cost by routing the input to a small subset of experts with a router. While MoEs are promising in terms of model performance and inference efficiency, they require careful design and tuning of the router. A practical router should either enable a balanced workload for experts to avoid downgrading the model performance or reduce the communication overhead to guarantee that training finishes in a reasonable time \cite{lepikhin2020gshard,fedus2021switch,lewis2021base}. For example, Switch Transformer \cite{fedus2021switch} was introduced to train a model with 1.6 trillion parameters by simplifying MoE to route each token to only a single expert and using an auxiliary loss to improve the workload balance during the training. 

Despite the success of the aforementioned literature in pre-training giant models, they have two main disadvantages. First, they rely on an All2All communication collective for both intra-node and inter-node data exchange, which have heterogeneous bandwidths that differ by a large gap (e.g., in AWS EC2 P4d, the peak bandwidth of EFA is 400Gbps, while the aggregated bandwidth of NVSwitch inside a node is 600GB/s), hence downgrading the communication efficiency. Second, computing a balanced router is increasingly more expensive as more experts are used. In Switch Transformer, a large number of experts cause congestion, generate network hotspots, and adversely affect performance. 

\begin{figure*}[h!]
    \centering
    \makebox[\textwidth][c]{\includegraphics[width=\textwidth]{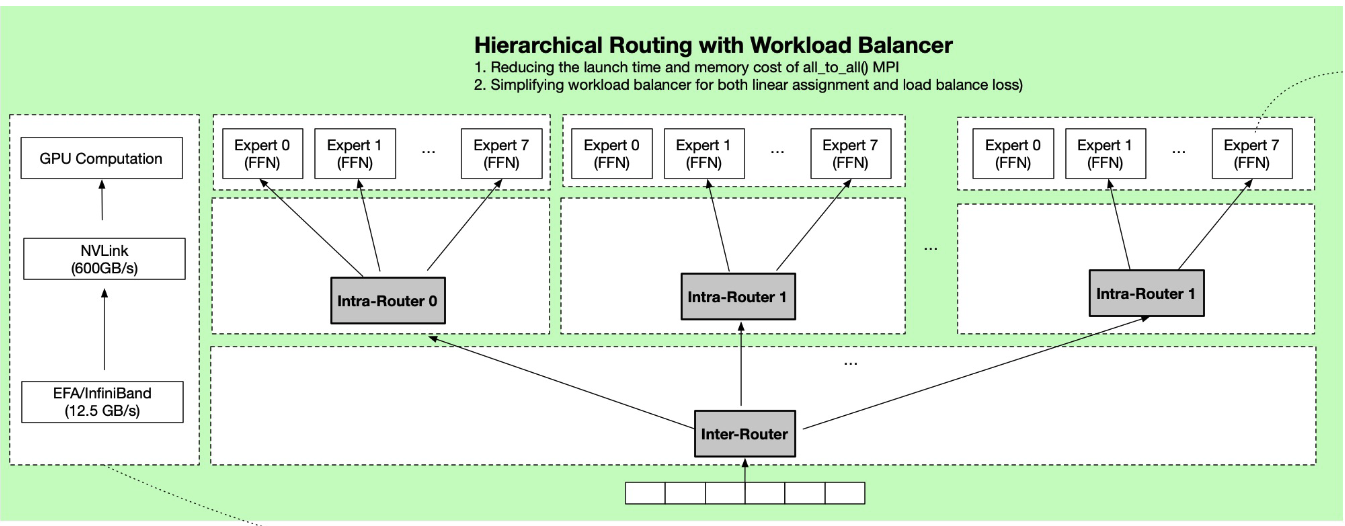}}
    \caption{Overview of \ours\ $(N = m \times n)$. Bi-level routing divides token dispatch into two stages, where in the first stage each token is routed to a node via an inter-node router and then gets assigned to a GPU by an intra-node router. This allows us to better utilize heterogeneous bandwidths to achieve greater communication efficiency. }
    \label{fig:bimoe}
\end{figure*}

To tackle the above bottlenecks, we introduce bi-level routing to scale up MoE with more efficient routing, dubbed as \ours. The system overview is illustrated in Figure~\ref{fig:bimoe}. We divide the experts into two levels based on the mesh topology. All the experts within a node are considered a group. Each token is first routed to a node and then gets dispatched to a particular GPU within the node. In this way, inter-node network congestion is dramatically reduced. Moreover, the launch overhead on a node for All2All communication is reduced from $\mathcal{O}(mn)$ to $\mathcal{O}(m + n)$, where $m$ and $n$ denote the number of GPUs per node and the total number of nodes, respectively. Similarly, the time complexity of routing is reduced from $\mathcal{O}(mnTd)$ to $\mathcal{O}(max(n, m)Td)$, where $T$ is the total number of tokens and $d$ is the model hidden size. 

In the experiments, we demonstrate that the proposed \ours\ improves throughput over Switch Transformer by 2.5$\times$ for training a 3.7B parameter model on 128 GPUs while being able to maintain the same convergence speed. The scalability analyses show that \ours\ achieves significantly better weak and strong scaling efficiencies than Switch Transformer, and maintains a good performance advantage when we increase the model size from 3.7B to 48B. The profiling of a single MoE layer confirms our motivation that All2All communication is the major bottleneck in MoE and the proposed bi-level routing significantly mitigates the communication overhead.

\section{Related Works}
\label{sec:related_works}
Mixture of Experts (MoE), in the context of modern deep learning architectures, was proven effective in \cite{shazeer2017outrageously}.
In this work, a MoE layer was applied between stacked LSTM \cite{hochreiter1997long} layers, and tokens were separately routed to different subsets of experts. In our work, we consider a hybrid of data + expert parallelism, where each worker holds a single expert for each MoE layer, and the number of experts scales with the number of workers, i.e., $N=nm$ with $m$ and $n$ denoting the number of GPUs per node and total number of nodes. For the token assignment, the router is equipped with a variable $W_r \in \mathcal{R}^{N\times d}$ and produces logits $r(x) = W_rx$, where $x \in \mathcal{R}^d$ is the token hidden vector. The logits are normalized via a softmax to construct the probability of selecting each expert. The routing probability for expert $e$ is given by
\begin{eqnarray}
p_e(x) = \frac{r_e(x)}{\sum_{i=1}^Nr_i(x)}.
\label{eq:moe_probability}
\end{eqnarray}
The top-k experts are then selected for processing the given token. Denote the set of chosen experts by $\mathcal{I}$. The output of the top-k experts is given by
\begin{eqnarray}
y(x) = \sum_{e \in \mathcal{I}}p_e(x)E_e(x),
\end{eqnarray}
where $E_e(\cdot)$ is the sub-model (e.g., multi-layer perceptron) for expert $e$. In this way, the number of model parameters increases linearly with respect to the number of experts only with a small amount of extra computational cost for routing. MoE offers state-of-the-art performance in language modeling and machine translation benchmarks. The routing step has a total complexity of $\mathcal{O}(kmnTd)$, where $T$ is the total number of tokens.  
The MoE layer was reintroduced into the Transformer architecture by the Mesh Tensorflow library \cite{shazeer2018mesh} where MoE layers were introduced as a substitute for the Feed-forward Network (FFN) layers in Transformers \cite{vaswani2017attention}. However, there were no accompanying NLP results.
With recent advances in machine learning infrastructure, GShard \cite{lepikhin2020gshard}, which extended the XLA compiler, uses MoE to dramatically improve machine translation across 100 languages.
In \cite{fan2021beyond}, a different deterministic MoE strategy is adopted to split the model parameters into non-overlapping groups of languages. Switch Transformer \cite{fedus2021switch} simplifies the routing process and only selects a single expert for each token. BASE \cite{lewis2021base} is another MoE variant that stacks multiple FFN layers as a single expert and inserts them into a standard Transformer architecture. This significantly increases the inference time compared with a vanilla Transformer.  

Our proposed method introduces bi-level routing to better leverage heterogeneous communication bandwidth, and utilize two additive losses for load balancing, achieving a large speedup. A concurrent work \cite{rajbhandari2022deepspeed} independently
considers intra-node all-to-all first followed by an inter-node all-to-all. We would like to emphasize that this work does not change routing mechanisms and only exploits hierarchical all-to-all for \textit{inference}. 

\section{Method}
\label{sec:formulation}

\subsection{Background: Bottleneck of MoE in Scaling to a Large Number of GPUs}
\label{sec:bg}

\begin{figure*}[h!]
\centering
\begin{minipage}{.60\textwidth}
\begin{python}
for (int r = 0; r < numranks; r++) {
    if (count != 0) {
        ncclSend(sendbuff + r*rankdiff, 
        count, type, r, comm, stream);
        
        ncclRecv(recvbuff + r*rankdiff, 
        count, type, r, comm, stream);
    }
}
\end{python}
\caption{\textcolor{black}{The implementation of All2All Communication in NCCL (torch/csrc/cuda/nccl.cpp)}}
\label{fig:code_all2all_nccl}
\end{minipage}
\hspace{0.2cm}
\begin{minipage}{.35\textwidth}
\centering
 {\includegraphics[width=\columnwidth]{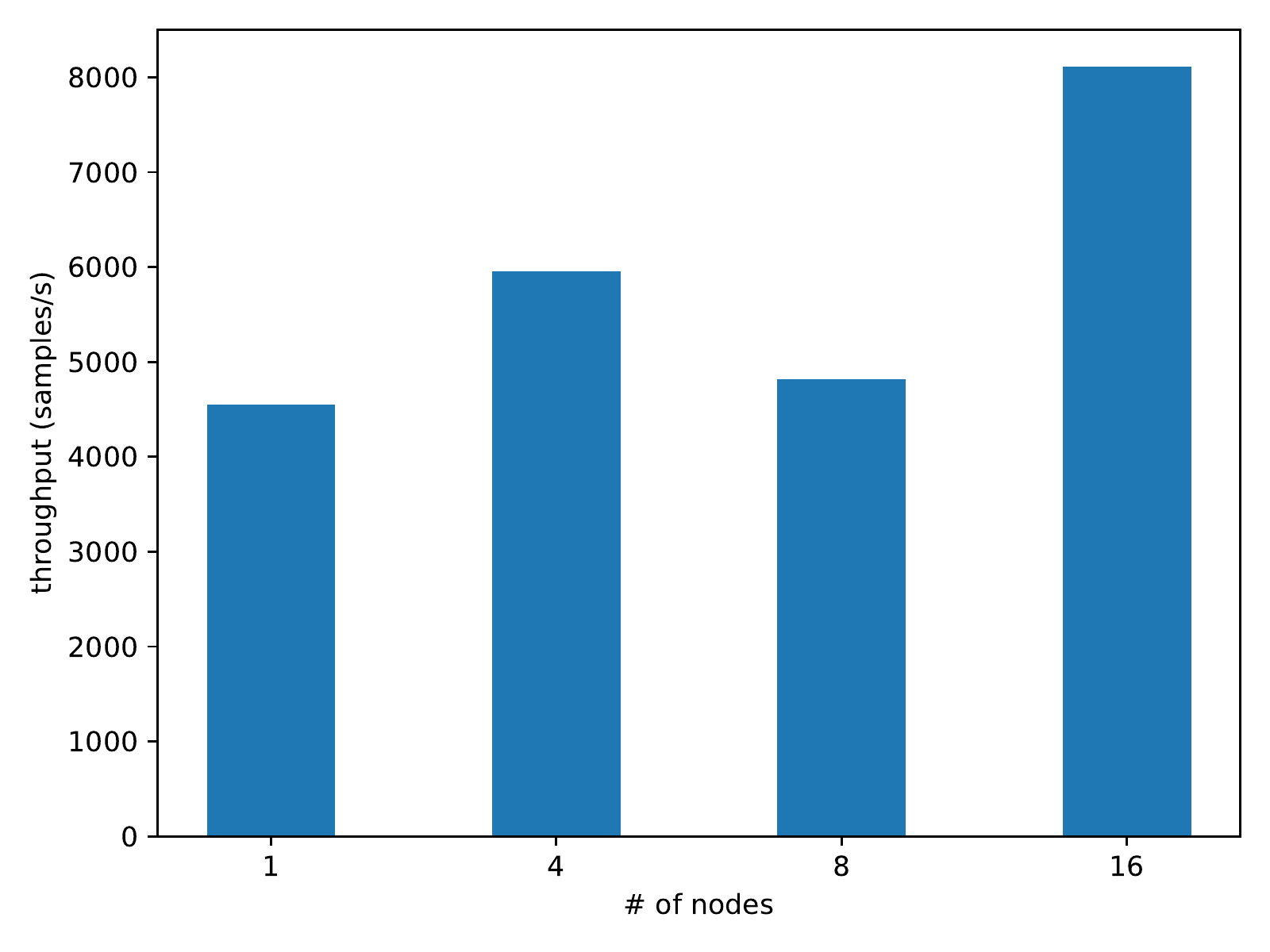}}
\captionof{figure}{Throughput results when scaling Switch Transformer to a large number of GPUs}
\label{fig:scale_all2all_nccl}
\end{minipage}

\end{figure*}

MoE heavily relies on the performance of All2All. Depending on the network topology, different All2All algorithms result in different communication costs, latency, and network congestion behaviors. Suppose that there are $N$ workers in total. For a ring topology, All2All has a quadratic communication cost and linear latency on $N$ while communication cost and latency are reduced to $\mathcal{O}(N^{3/2})$ and $\mathcal{O}(N^{1/2})$, respectively for mesh topology such as TPUs \cite{kumar1994introduction}. Regardless of the underlying topology of the network, another trivial approach is to send all the messages asynchronously into the network as illustrated in Figure~\ref{fig:code_all2all_nccl}. This naive algorithm implements pairwise one-to-one routings \footnote{\url{https://github.com/pytorch/pytorch/blob/2b7943c47c8561a46103488b0fe9a592b87dc5bb/torch/csrc/cuda/nccl.cpp\#L637}} and suffers from network congestion because of the bisection width of the network \cite{hambrusch1995communication}. As in Figure~\ref{fig:scale_all2all_nccl}, Switch Transformer has very poor scaling efficiency when the number of nodes is increased from 1 to 16 (8 to 128 GPUs). The throughput on 8 nodes are even worse than that on 4 nodes.

\subsection{Efficient MoE Layer via Bi-level Routing}


To tackle the above bottlenecks, we introduce bi-level routing to scale up MoE with more efficient routing, dubbed as \ours.

\subsubsection{Model Architecture: Orchestration of Inter-node All2All and Intra-node All2All}
\label{sec:model_architecture}

\begin{figure*}[h!]
  \centering
    \centering
     {\includegraphics[scale=0.4]{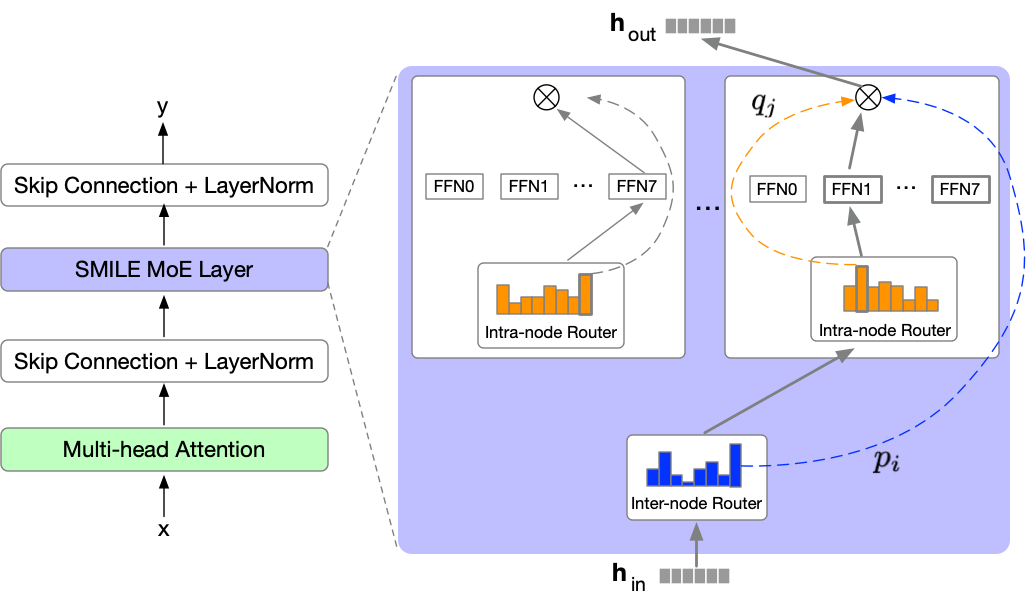}}
    \captionof{figure}{Illustration of \ours\ Layer (m = 8, n = 8)}
    \label{fig:moe_illustration}
\end{figure*}
        
To account for the heterogeneous and hierarchical nature of the inter-connection network, we divide the experts into two-level All2All operation. All the experts within a node are grouped together. As shown in Figure \ref{fig:moe_illustration}, when a token is ready for the dispatch, it is first routed to a node via an inter-node router (blue) and is then assigned to a GPU via an intra-node router (orange). 

The proposed bi-level routing reduces the launch time on a node for All2All from $\mathcal{O}(mn)$ to $\mathcal{O}(m + n)$, where $m$ and $n$ are the number of GPUs per node and the total number of nodes, respectively. In terms of the communication efficiency, bi-level routing parallelizes multiple All2All collectives, which minimizes network interference between flows and significantly reduces inter-node network congestion compared to the naive algorithm implemented in NCCL (Ref. Figure \ref{fig:code_all2all_nccl}).

Bi-level routing also simplifies router optimization where the size of routing problem in Switch Transformer (Equation \ref{eq:moe_probability}) is decreased from $mn$ to $max(n, m)$. In particular, the complexity of routing in Switch Transformer is reduced from $\mathcal{O}(mnTd)$ to $\mathcal{O}(\max(n, m)Td)$.

The dash curve in Figure \ref{fig:moe_illustration} also illustrates the way we calculate the output of the top-1 expert:
\begin{eqnarray}
h_{out} = p_i(h_{in})q_{j}(h_{in})E_{i, j}(h_{in}),
\end{eqnarray}
where $p_i(h_{in})$ and $q_{j}(h_{in})$ are the top-1 routing probability (generated by Equation \ref{eq:moe_probability}) allocated for node $i$ and local expert $j$, respectively, and $E_{i, j}$ is the $j$-th expert on node $i$. Both the inter-node and intra-node routers have tied parameters across all the workers, ensuring that the results do not change when an incoming example is processed by a different worker. Compared to a single router, bi-level routers also reduce the total number of router parameters from $\mathcal{O}(mn)$ to $\mathcal{O}(m + n)$.

\subsubsection{Additive Load Balancing Loss}

Different from the one-hop load balancing loss in previous works \cite{shazeer2017outrageously,shazeer2018mesh,lepikhin2020gshard,fedus2021switch}, we use an additive load balancing (LB) loss for bi-level routing. Given $N$ experts indexed by $i=1$ to $N$, we decouple it into $n \times m$, where $n$ is the number of GPU nodes and $m$ is the number of GPUs on a single node (e.g., 8 GPUs is a common configuration for a GPU node). For a batch $\mathcal{B}$ with $T$ tokens, the additive load balancing loss has two components, either of which is computed as the scaled dot-product between dispatch fraction and routing probability vectors:

\begin{equation}  \label{eq:bi_level_lb_loss}
\text{loss}_{\text{lb}} = \underbrace{\alpha \cdot n \cdot \sum_{i=1}^{n} f_i \cdot P_i}_{\text{inter-node LB loss}} + \underbrace{\beta \cdot m \cdot \sum_{j=1}^{m} f_j \cdot Q_j}_{\text{intra-node LB loss}},
\end{equation}

\noindent where $f_i$ and $f_j$ are the fraction of tokens dispatched to node $i$ and local expert $j$, respectively: $f_i = \frac{1}{T}\sum_{x \in \mathcal{B}} 1 \{\texttt{argmax} p(x) = i\} $, $f_j = \frac{1}{T}\sum_{x \in \mathcal{B}} 1 \{\texttt{argmax} q(x) = j\} $; $P_i$ and $P_j$ are the fraction of the router probability allocated for node $i$ and expert $j$ respectively;
$P_i = \frac{1}{T}\sum_{x \in \mathcal{B}} p_i(x)$, $Q_j = \frac{1}{T}\sum_{x \in \mathcal{B}} q_j(x)$; hyper-parameter $\alpha$ and $\beta$ are multiplicative coefficients. The minimum is attained under uniform inter-node and intra-node routing, i.e., $\min \text{loss}_{\text{lb}} = \alpha \cdot n \cdot \sum_{i=1}^{n} 1/n \cdot 1/n + \beta \cdot m \cdot \sum_{j=1}^{m} 1/m \cdot 1/m = \alpha + \beta$. In practice, we simply use $\alpha = \beta$.

To compute the total model loss during training, we sum up $\text{loss}_{\text{lb}}$ in all \ours\ layers as the auxiliary loss:
\begin{equation}  \label{eq:total_loss}
\text{loss}_{\text{total}} = \text{loss}_{\text{train}} + \sum_{l=1}^{L} {\text{loss}_{\text{lb}}^{l}},
\end{equation}
where $L$ denotes the total number of \ours\ layers and $\text{loss}_{\text{lb}}^{l}$ is the load balancing loss in $l$-th \ours\ layer.

\subsubsection{Bi-level Process Group Management}

\begin{figure*}[h!]
  \centering
    \centering
     {\includegraphics[width=\columnwidth]{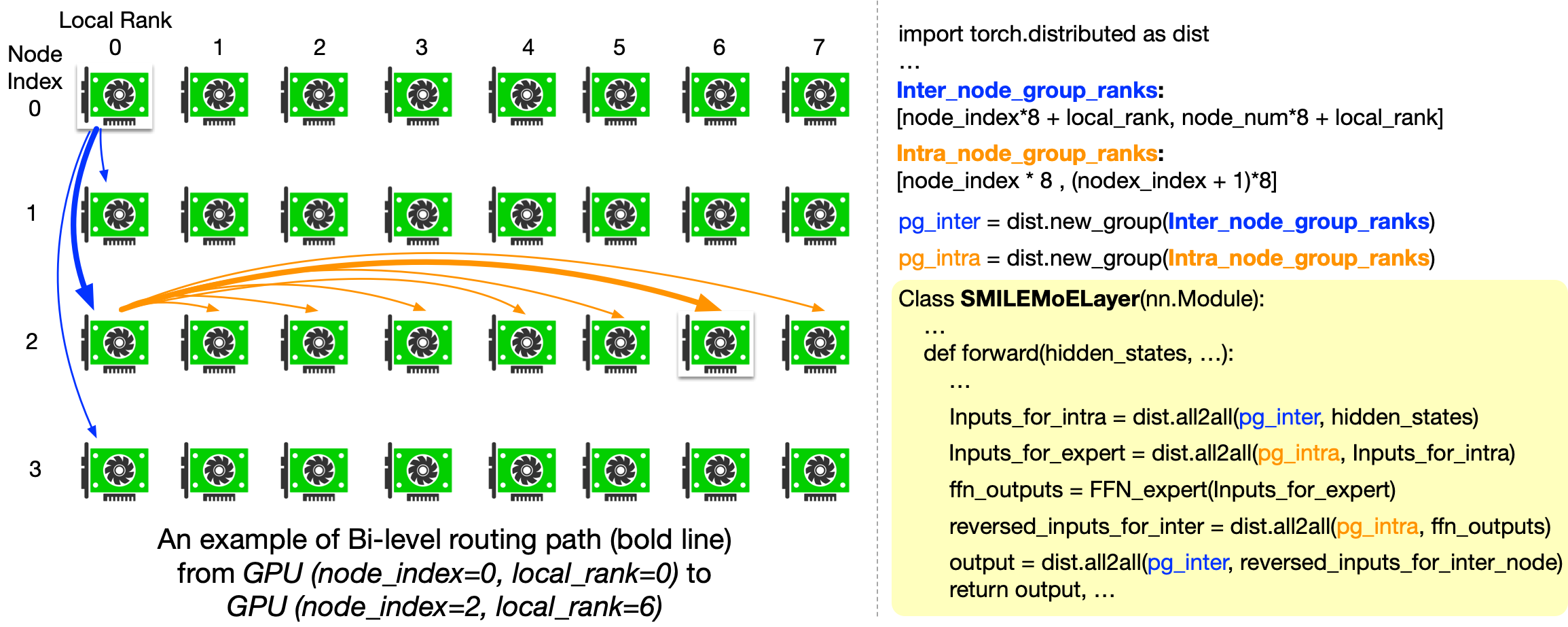}}
    \captionof{figure}{Bi-level Process Group Management and its Pseudocode}
    \label{fig:process_group_management}
\end{figure*}

The system implementation of the \ours\ layer requires two levels of distributed process management. The first level process group handles node-level All2All, and the second level manages All2All between intra-node GPU processes. In addition, these two groups of processes should be connected to complete the bi-level routing without mutual interference. The left side of Figure~\ref{fig:process_group_management} shows the process of cooperation between the first-level inter-node process group and the second-level intra-node process group to complete bi-level routing. Based on such requirements, we propose a process management mechanism based on PyTorch \texttt{dist.new\_group} API. As shown on the right, for each GPU process, we create an inter-node process group and an intra-node process group, where the process ranks in the group are shown in blue text and orange text, respectively. Based on this process group management, when performing the All2All operation for the BiMoE Layer, we only need to specify the \texttt{inter\_node\_process\_group} instance and \texttt{intra\_node\_process\_group} instance according to local rank. This method greatly simplifies the management of the process group so that the MoE layer itself does not need to care about the system implementation details. The right side of the figure also shows the process of four sequential All2All operations. Two additional All2All operations are required because of the reversed routing for the consecutive attention layer.





\section{Experiments}

\subsection{Experimental Setup}
\paragraph{Task and Dataset.} We evaluate \ours\ on NLP pre-training tasks with large Transformer models. 
We use a masked language modeling task \cite{taylor1953cloze,fedus2018maskgan,devlin2018bert} where the model is trained to predict missing tokens. We evaluate the performance of \ours\ by pre-training on the “Colossal Clean Crawled Corpus” (C4), a collection of English-language text sourced from the public Common Crawl web scrape. It includes heuristics to extract only natural language (as opposed to boilerplate and other gibberish) in addition to extensive deduplication \cite {raffel2019exploring}. The C4 dataset is obtained from the curated version hosted by Hugging Face Dataset \footnote{\url{https://huggingface.co/datasets/c4}}. It has 129 billion tokens (words) in the training dataset and 129 million tokens (words) in the validation dataset. For the parallel training on large number of GPUs, we split the training dataset into 32768 (1024 x 24) files, and validation dataset into 256 files. We use the same vocabulary as the original T5 (11B) model \footnote{\url{https://github.com/huggingface/transformers/blob/main/src/transformers/models/t5/tokenization\_t5.py}} (vocabulary size is 32128).

\paragraph{Model Architecture.} We compare \ours\ to Switch Transformer to demonstrate the efficiency of bi-level routing. The proposed method can also be used in conjunction with other MoE models such as GShard \cite{lepikhin2020gshard} and BASE \cite{lewis2021base}.  For fair comparison, Switch Transformer and \ours\ use the same BERT-like architectures (a stack of many standard Transformer layers) but replaces the every other feed forward network (FFN) layer in Transformer with a MoE (mixture of experts) layer. In each Transformer layer, the MoE layer follows after a multi-head attention layer, and they are enhanced with a skip connection followed by a LayerNorm operation afterwards. The activation function in attention and FFN layer is set to GELU with a dropout rate 0.1. In \ours\, the routing architecture is defined according to the design in Section \ref{sec:model_architecture}. For different model sizes, we only change the number of hidden layers, hidden size, and intermediate size (details are introduced in Section \ref{exp:model_size}). 

\paragraph{Hardware.} We run experiments on state-of-the-arts hardware in AWS with advanced supports for computing, communication, storage: \textit{1. GPU accelerators}: we evaluate all baselines and \ours\ on AWS P4d nodes \footnote{\url{https://aws.amazon.com/ec2/instance-types/p4/}}. Each node is equipped with 8 NVIDIA A100 GPUs. We scale up to 16 nodes to evaluate the  scalability;
\textit{2. High bandwidth Communicator}: We utilize AWS EFA (Elastic Fabric Adapter) \footnote{\url{https://aws.amazon.com/hpc/efa/}} for 400 Gbps high bandwidth inter-node networking. Compared to commonly used NVIDIA InfiniBand, EFA's custom-built operating system (OS) bypass hardware interface enhances the performance of inter-instance communications. Our experiments show that even in such an high bandwidth setting, the All2All communication is still a bottleneck in MoE models (e.g., Switch Transformer);
\textit{3. Networked File System}: To boost the performance of accessing the data files in a distributed manner, we use AWS FSx \footnote{\url{https://aws.amazon.com/fsx/}} with SSD support. The total storage cost for C4 dataset and all source code is around 800G.

\paragraph{Training Hyper-parameters.} We train MoE models with the LAMB optimizer \cite{you2019large}, where the learning rates are tuned in the range $\{0.0001, 0.0003, 0.001, 0.003\}$, the weight decay is fixed to 0.01, and $\epsilon$ is set to 1e-6. We clip gradients if their l2 norm exceeds 1.0. As a common practice to reduce the GPU memory cost in LAMB optimizer, we also enable half precision (fp16). We use a sequence length of 128. Unless otherwise specified, we fix overall training batch size to 16384 and micro batch size to 128, where micro batch size refers to the batch size per GPU per micro step and total\_batch\_size = micro\_batch\_size * num\_micro\_steps. Gradient accumulation is adopted when the number of micro steps is larger than 1. We use 128 because it is the maximum
size that can be used under GPU memory
constraints with our hardware configuration. For scalability analysis, we scale the number of nodes from 1 node (8 GPUs) to 16 nodes (128 GPUs).

\paragraph{Implementation.} Our source code is well-maintained as a Python pip package.
We implement our algorithm with the integration of the PyTorch DDP and DeepSpeed frameworks. The process group management introduced in is handled by PyTorch DDP \cite{li2020pytorch} grouping APIs. We use LAMB optimizer implemented by DeepSpeed \cite{rasley2020deepspeed}. For GPU memory-efficient training of large dense model, we reuse a few techniques supported by DeepSpeed, including ZERO optimization \cite{rajbhandari2020zero}, activation checkpointing, and half precision (fp16). To analyze the fine-grained time breakdown for communication and computation in MoE layer, we use PyTorch Profiler.  
Our data loader for C4 dataset is customized with the pre-fetching mechanism for efficient distributed loading.




\subsection{Comparison with BERT and Switch Transformer}


\begin{figure*}[h!]
  \centering
  \begin{minipage}{.48\textwidth}
    \centering
     {\includegraphics[width=\columnwidth]{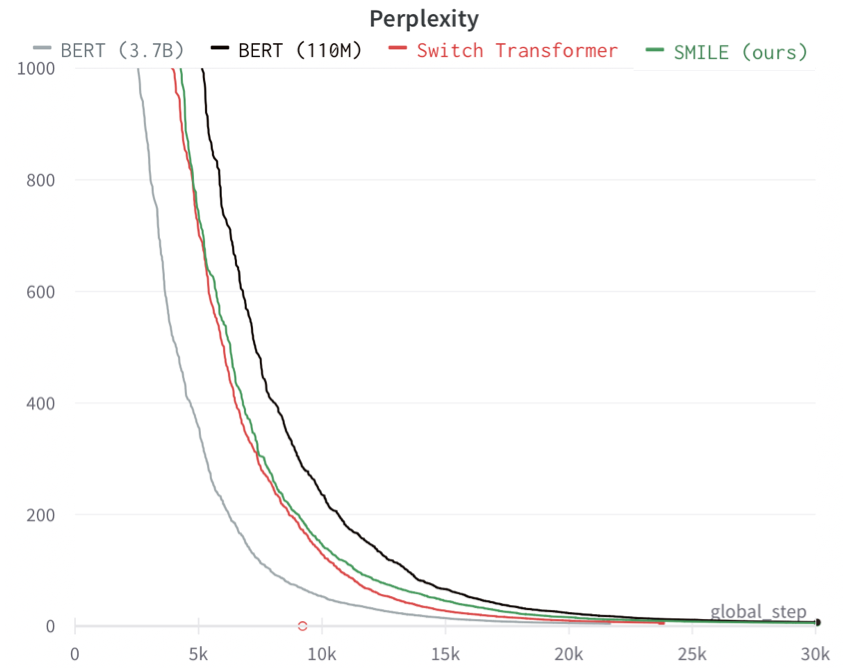}}
    \captionof{figure}{The curve of iteration-to-perplexity}
    \label{fig:iteration_to_ppl}
  \end{minipage}
  \begin{minipage}{.48\textwidth}
    \centering
      {\includegraphics[width=\columnwidth]{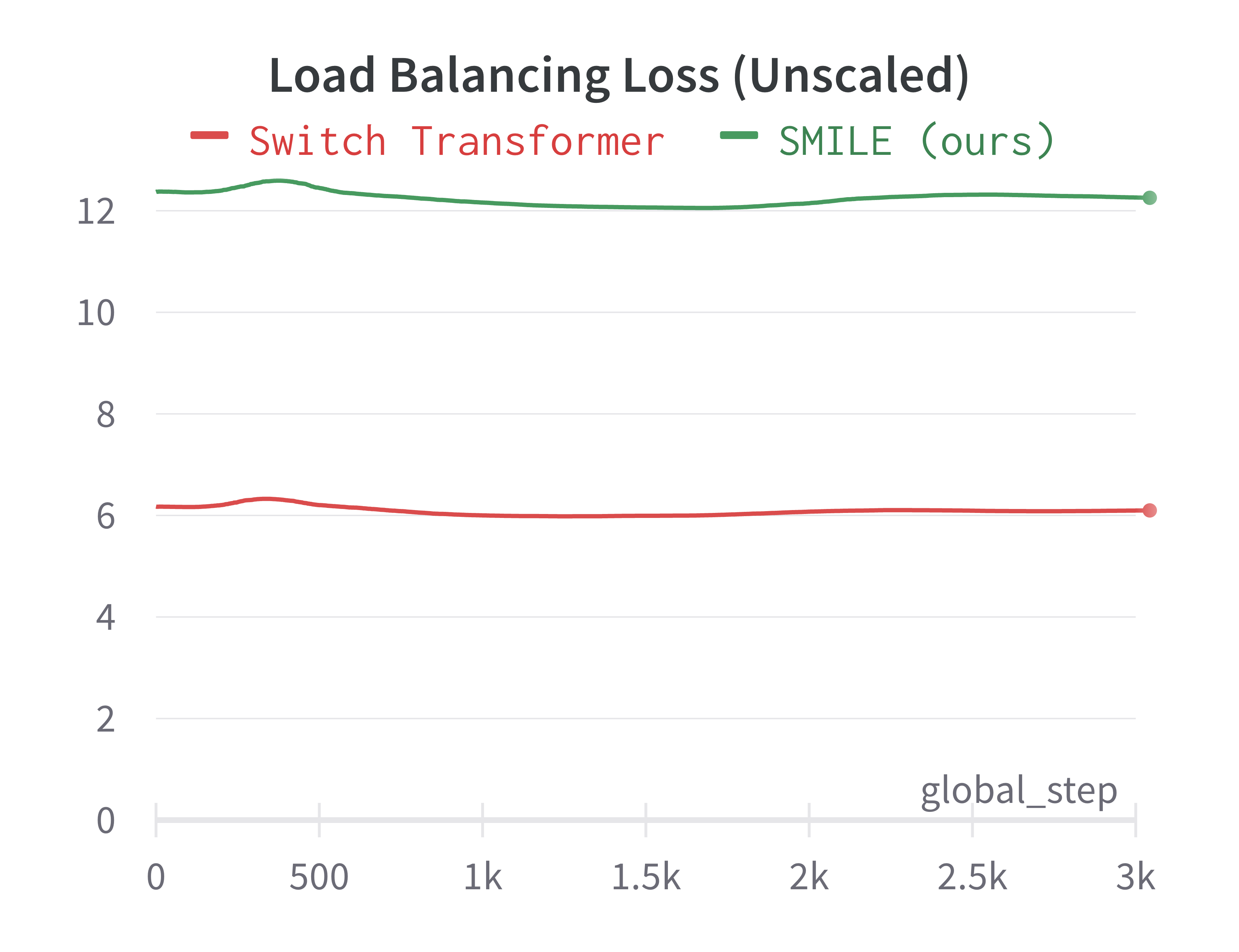}}
    \captionof{figure}{Unscaled load balancing loss.}
    \label{fig:lb_loss}
  \end{minipage}
\end{figure*}






\paragraph{Accurate and Faster Training.} In addition to Switch Transformer, we compare \ours\ with BERT (110M) and BERT (3.7B) baselines which have the same model floating point operations (FLOPs) and number of parameters as Switch Transformer and \ours, respectively. We use $\alpha=0.01$ for Switch Transformer and $\alpha=\beta=0.005$ (introduced in Equation (\ref{eq:bi_level_lb_loss})) for \ours\ in our experiments, and set the capacity factor for routing as 2.0. We replace every other shared feed-forward layer in the Transformer architecture with a MoE (Resp. \ours) layer.

\begin{wraptable}{r}{0.46\textwidth}
\vspace{-5mm}
    \centering
    \caption{Throughput (samples/second)}
    \label{tab:model_throughputs}
   \begin{tabular}{c|c}
   \toprule
       Model  & Throughput \\
       \midrule
       BERT (110M)  & 93282\\
       BERT (3.7B)  & 5114\\
       Switch Transformer  & 8112\\
       \ours\  & 20011\\
       \bottomrule
    \end{tabular}
    \label{tab:model_throughputs}
\end{wraptable}
From Figures~\ref{fig:iteration_to_ppl}, ~\ref{fig:lb_loss} and Table~\ref{tab:model_throughputs}, we have four important observations. First, \ours\ has the same convergence behaviour as Switch Transformer, and it converges faster than BERT (110M). Second, both \ours\ and Switch Transformer converge slower than BERT (3.7B), which is expected since the MoE models trade-off convergence for greater computational efficiency. Third, both Switch Transformer and \ours\ are slower than BERT (110M), indicating that routing is the major bottleneck in the MoE models. And, \ours\ runs 2.5x and 3.9x faster than Switch Transformer and BERT (3.7B), respectively. This proves that bi-level routing is effective in reducing the overhead of standard MoE layers. Lastly, \ours\ achieves the twice unscaled balancing loss of that of Switch Transformer, which is expected since \ours\ has two additive losses. When scaled with $\alpha$, two curves will roughly overlap with each other.


Next, we provide fine-grained performance analysis and ablation studies to justify the necessity of bi-level routing.

\subsection{Scalability}
\subsubsection{Scalability on High Bandwidth (400 Gbps) Inter-node Communication}
 
\begin{figure*}[h!]
\setcounter{subfigure}{0}
\subfigure[\label{fig:fig1a} Weak Scaling]
{{\includegraphics[width=0.50\textwidth]{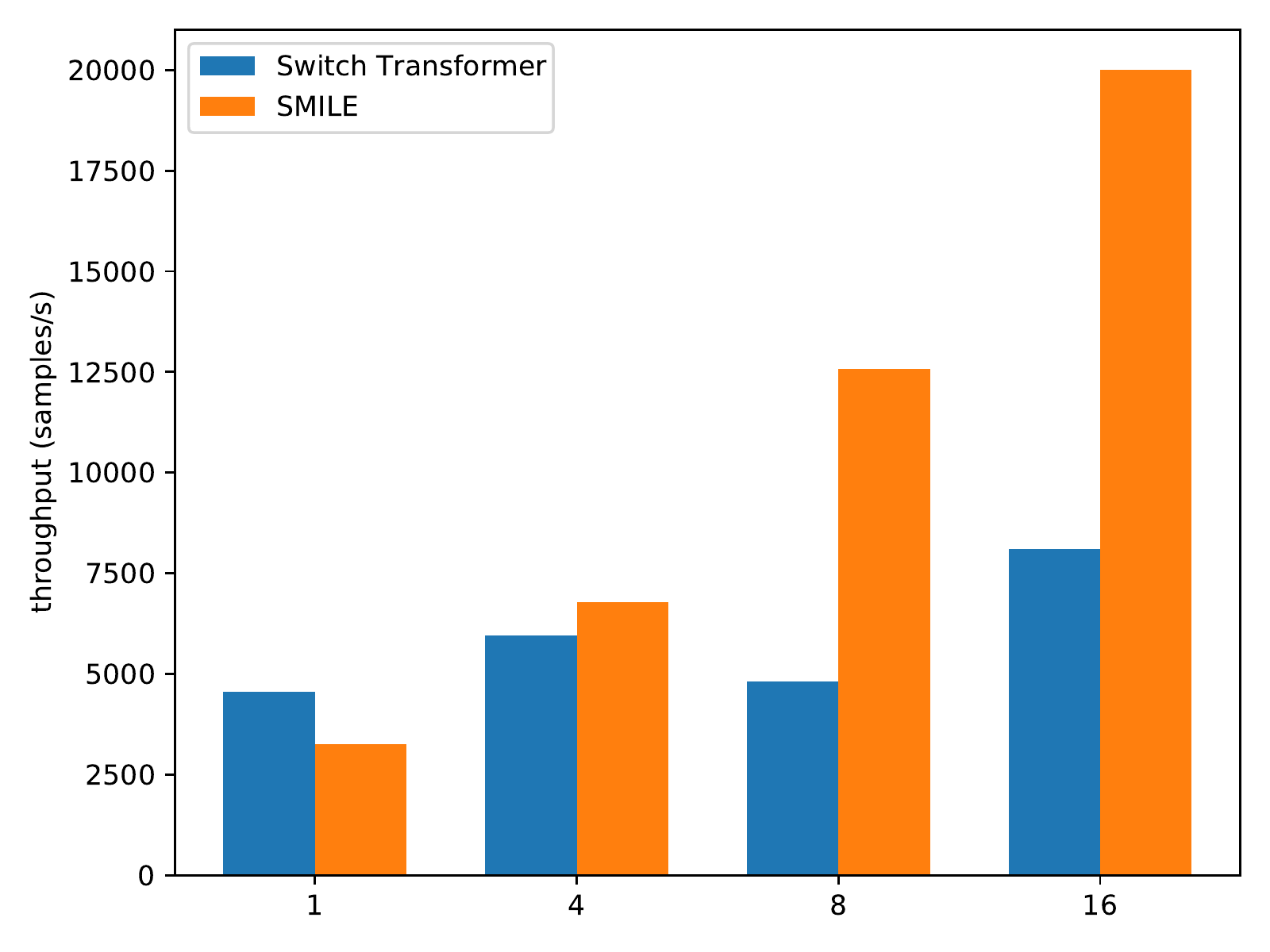}}}
\subfigure[Strong Scaling]
{{\includegraphics[width=0.50\textwidth]{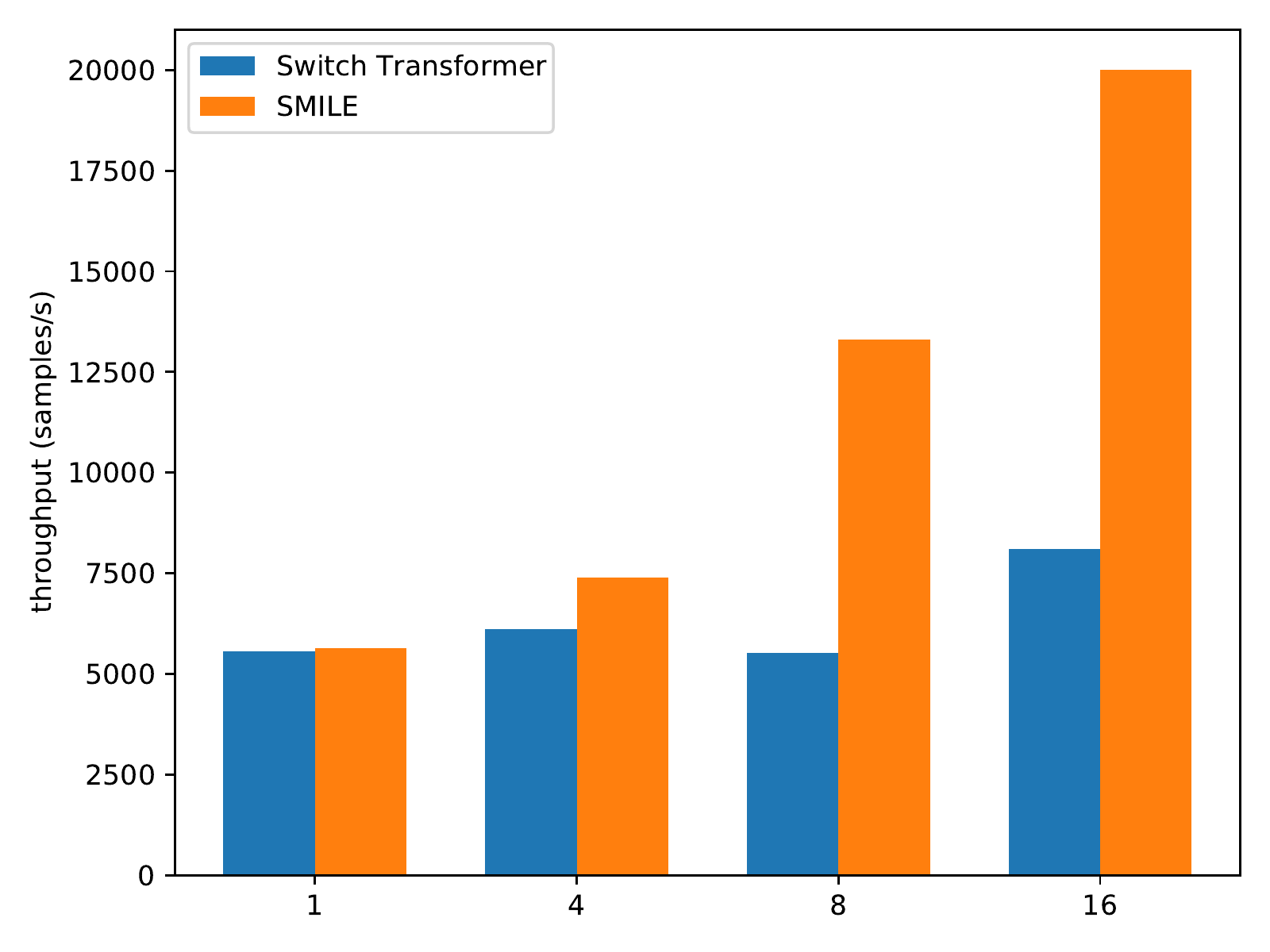}}}
\caption{\textcolor{black}{Switch Transformer vs. \ours}. The number of nodes is increased from 1 to 16.}
\label{fig:performance_node_scaling}
 \end{figure*}
 
We compared the throughput (samples per second) between \ours\ and Switch Transformers when scaling the number of GPU nodes (each node has 8 GPUs) from 1 to 16 in high bandwidth. Both weak scaling and strong scaling are evaluated. In weak scaling, the global batch size is adjusted with the number of GPUs, while in strong scaling, both the global batch size and the micro batch size per GPU are fixed (the number of gradient accumulation steps decreases when the node number is scaling up). From the results in Figure \ref{fig:performance_node_scaling}, we have the following observations. 

1. MoE overhead in Switch Transformer (All2All Communication) is non-trivial even in a large bandwidth supported by advanced communication adaptor (AWS EFA). Its scaling efficiency is far below the linear scaling, which can be explained by the additional inter-node communication cost that drags down the performance; what is even worse is that the final throughput on 16 nodes is not notably better than that on a single node and 8 nodes has worse throughput than 4 nodes.

2. Compared to Switch Transformer, \ours\ scales up much better from 1 node to 16 nodes. The throughputs on 16 nodes are 7.7x and 4x higher than those on 1 node with weak and strong scaling, respectively. Moreover, different from Switch Transformer, when scaling from 4 nodes to 8 nodes, the throughput still increases. We observe worse performance of \ours\ on 1 node with weak scaling, which is due to additional overhead in the implementation. On a single node, we should directly use Switch Transformer.

Therefore, we conclude that bi-level routing is efficient in inter-node MoE scaling, and the scalability is largely improved when \ours\ is applied.

\subsubsection{Scalability w.r.t. Different Model Sizes}
\begin{table}[h!]
    \centering
    \caption{Comparison of Throughput between Switch Transformer and \ours\ (16 P4d nodes). We fix the total batch size to 16384 and vary the micro batch size depending on the model size and GPU memory.}
    \resizebox{\columnwidth}{!}{\begin{tabular}{cccc}
    \toprule
        \multirow{2}{*}{Model Size (128 Experts)} & \multirow{2}{*}{Model Configuration} &  \multicolumn{2}{c}{Throughput (samples/second)} \\
       \cmidrule(lr){3-4}
       &  & Switch Transformer & \ours \\
    \midrule  
    3.7B &  \makecell{
        micro\_batch\_size = 128 \\
    num\_layers = 12\\
    hidden\_size = 768\\
    intermediate\_size = 3072} & 8112 & 20011 ($2.47\times \uparrow$)  \\
    \midrule
    13B &  \makecell{
    micro\_batch\_size = 64 \\
    num\_layers = 24\\
    hidden\_size = 1024\\
    intermediate\_size = 4096} & 4001 & 6829 ($1.71\times \uparrow$) \\
    \midrule
    48B &  \makecell{
    micro\_batch\_size = 64 \\
    num\_layers = 36\\
    hidden\_size = 1600\\
    intermediate\_size = 6400} & 889 & 2223 ($2.50\times \uparrow$) \\
    \bottomrule
    \end{tabular}}
    \label{tab:modelsize_eval}
\end{table}

To understand the performance in different model sizes, we evaluate on three model configurations as shown in Table \ref{tab:modelsize_eval}. We conduct this experiment on 128 GPUs, and the number of experts is fixed to 128. The first two use BERT\_base and BERT\_large as the base dense models, while the third one is constructed by increasing the hidden size and model depth. This result demonstrates that \ours\ is not restricted to a specific model architecture, and still achieves 1.7-2.5 times faster training speed when the model size increases significantly.

\subsection{Understanding \ours: Time Breakdown and Performance Analysis}
\label{sec:breakdown}
To demystify the scalability benefit of \ours, we also did fine-grained performance analysis. It would be difficult and inaccurate by directly dissecting the performance of a single MoE layer from an end-to-end training pipeline, since there are other factors involved by the interaction between data parallelism (AllReduce) and MoE layer (All2All). As such, we develop a tiny model with only a single MoE layer and perform training with dummy data on the same GPU cluster with AWS EFA (16 P4d nodes)\footnote{Our source code also includes this evaluation framework}. By this way, we dissect the CUDA time cost for different phases in the MoE layer using PyTorch Profiler.

\begin{figure*}[h!]
    \centering
    \makebox[\textwidth][c]{\includegraphics[width=1\textwidth]{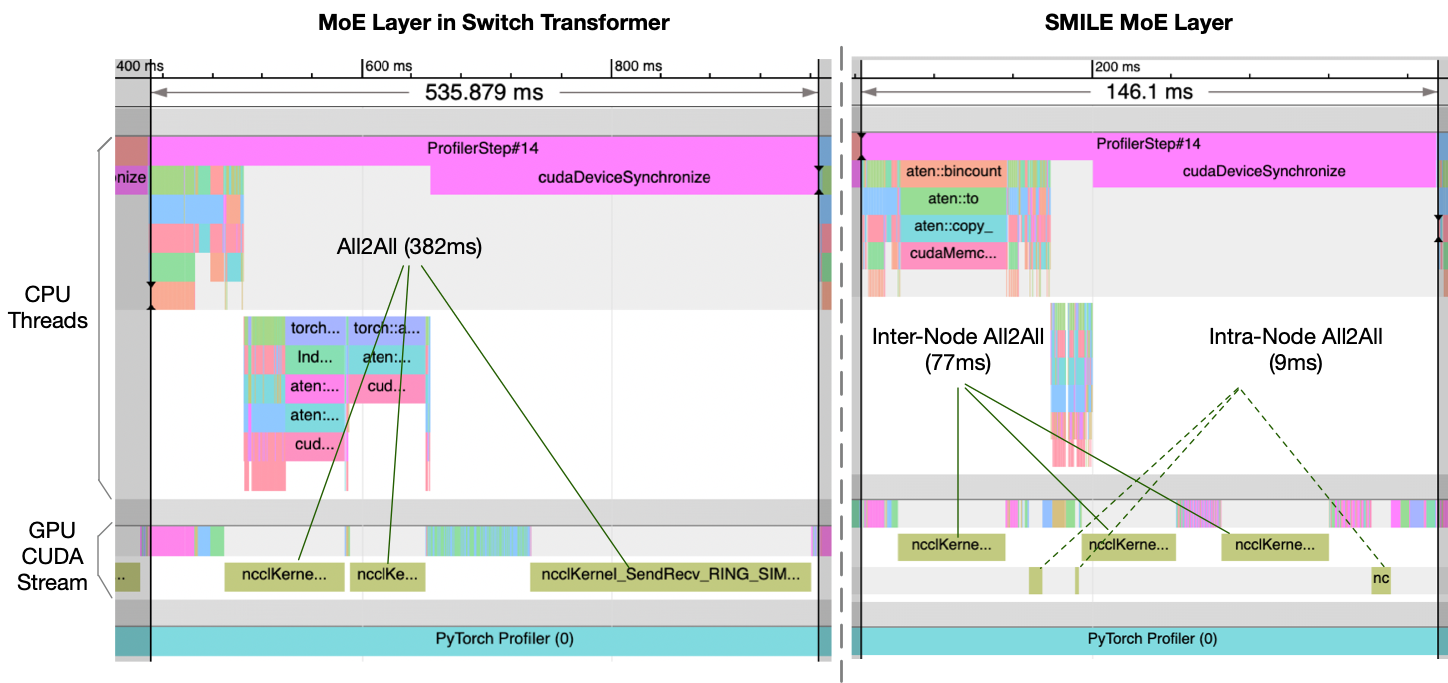}}
    \caption{\textcolor{black}{Dissecting the time cost in MoE layers (16 nodes; screenshotted via PyTorch Profiler)}}
    \label{fig:bimoe_analysis}
\end{figure*}

\begin{table}[h!]
    \centering
    \caption{Breakdown of the time cost per iteration (micro-batch FP) in MoE layers (16 P4d nodes)}
    \label{tab:performance_table}
    \resizebox{\columnwidth}{!}{\begin{tabular}{ccccc}
    \toprule
     & \multicolumn{2}{c}{Switch Transformer}& \multicolumn{2}{c}{\ours}  \\
     \midrule
     Total Time &  \multicolumn{2}{c}{535 ms} & \multicolumn{2}{c}{146 ms}  \\
     \midrule
    \multirow{2}{*}{All2All Time Cost} &\multicolumn{2}{c}{\multirow{2}{*}{\colorbox{lightgray}{382 ms}}} & Inter node & \colorbox{lightgray}{77 ms}   \\
    \cmidrule(lr){4-5}
    &  &  & Intra node & \colorbox{lightgray}{9 ms}  \\
     \midrule
     FFN Expert and Others (e.g., operations other than All2Alls)& \multicolumn{2}{c}{153  ms}  & \multicolumn{2}{c}{60 ms}  \\
     \midrule
      Ratio (All2All Time vs. Total Time)& \multicolumn{2}{c}{\colorbox{yellow}{71\%}}  & \multicolumn{2}{c}{\colorbox{yellow}{59\%}}  \\
      \bottomrule
    \end{tabular}
    }
\end{table}

The results are summarized in Figure \ref{fig:bimoe_analysis} and Table \ref{tab:performance_table}. In Figure \ref{fig:bimoe_analysis}, we mainly annotate the time cost for the All2All operations (due to two additional hops in routing for the reversed order, \ours\ has more All2Alls). The following observations provide us with clear evidence to support our motivation to design bi-level routing: (1) \ours\ can largely improve the overhead of a single MoE layer: bi-level routing running time (including EFA communication, NVSwitch communication, and GPU computation for expert networks) is 3.7 times less than one-hop routing across many nodes (146 ms vs. 535 ms); (2) All2All time cost in \ours\ is also 4.4 times smaller than that of Switch Transformer (382 ms vs. 86 ms), matching the analysis we have explained in Section \ref{sec:bg}; (3) Compared to the time cost on inter-node All2Alls (77 ms), the time cost on intra-node All2Alls (9 ms) is much smaller due to higher bandwidth (600GB/s vs. 50GB/s); (4) When applying \ours, the ratio (All2All Time vs. Total Time) is also reduced from 71\% to 59\%.

To understand more fine-grained details of the performance, we refer to the whole screenshot for visualizing performance results from PyTorch Profiler in Figure \ref{fig:per_switch} and \ref{fig:per_bi_moe}  in Appendix.

\section{Conclusion}
We propose a new routing algorithm and system for sparsely activated mixture-of-experts (MoE) layer. Specifically, we introduce \ours\ with bi-level routing that better leverages heterogeneous communication bandwidth. The bi-level routing significantly reduces network contention, launch overhead, and routing complexity. Our experiments demonstrate that the proposed \ours\ improves training throughput by 2.5$\times$ compared to SwitchTransformer on 128 GPUs without affecting convergence.

\bibliography{references}
\bibliographystyle{unsrt}

\appendix
\onecolumn
\appendix

\clearpage
\section{Appendix}

\subsection{Performance Profiling on Different MoE Layers}
\label{exp:moe_profiling}

\begin{figure*}[h!]
    \centering
    \makebox[\textwidth][c]{\includegraphics[width=\textwidth]{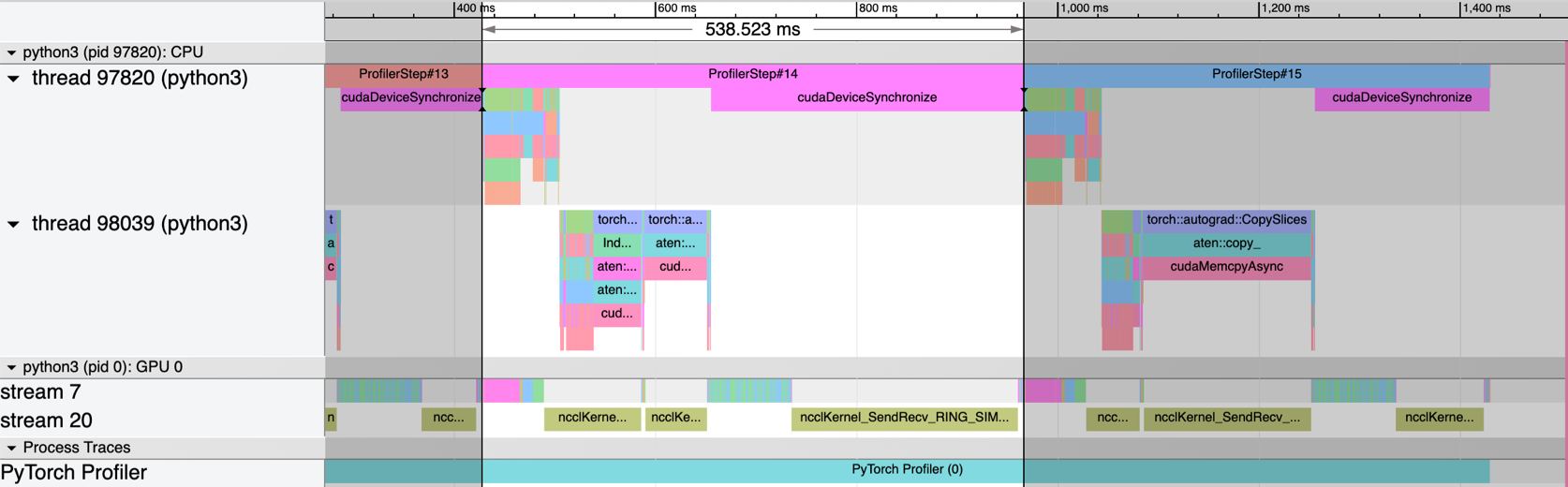}}
    \caption{\textcolor{black}{Switch Transformer MoE layer All2All time cost profiling (16 P4d nodes)}}
    \label{fig:per_switch}
\end{figure*}




\begin{figure*}[h!]
    \centering
    \makebox[\textwidth][c]{\includegraphics[width=\textwidth]{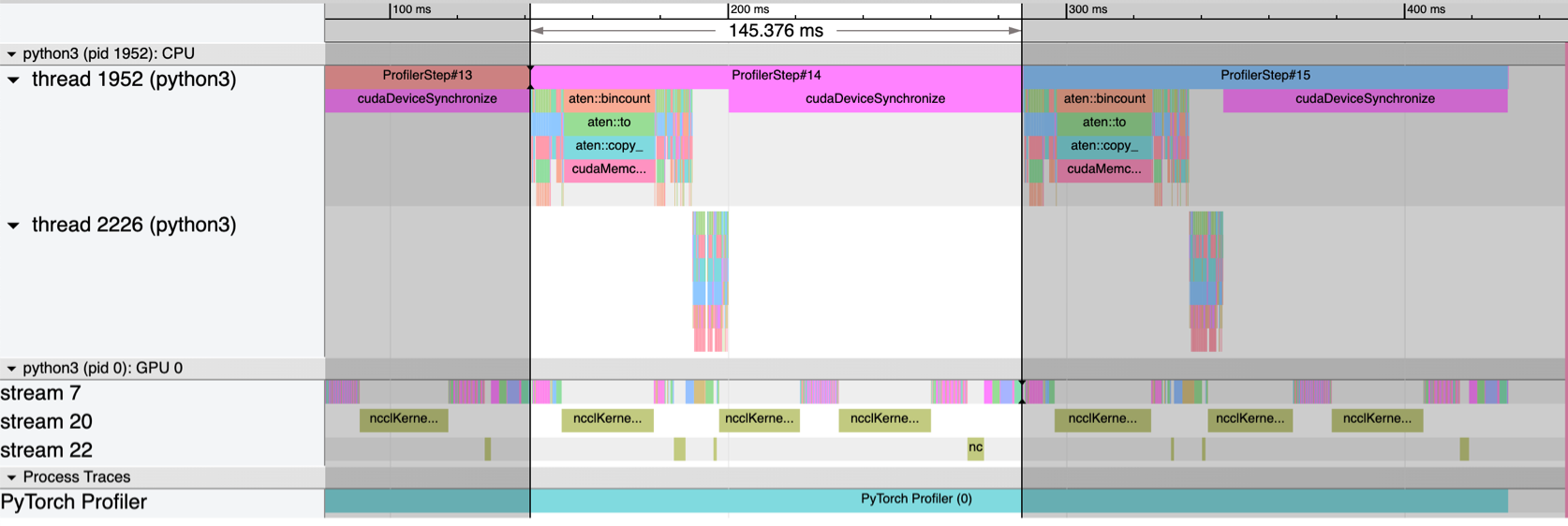}}
    \caption{\textcolor{black}{\ours\ layer All2All time cost profiling (16 P4d nodes)}}
    \label{fig:per_bi_moe}
\end{figure*}






\subsection{Another Angle to Understand the Overhead of All2All Communication}
\label{exp:model_size}

\begin{figure}[ht!]
    \centering
     \makebox[\linewidth][c]{\includegraphics[width=0.60\linewidth]{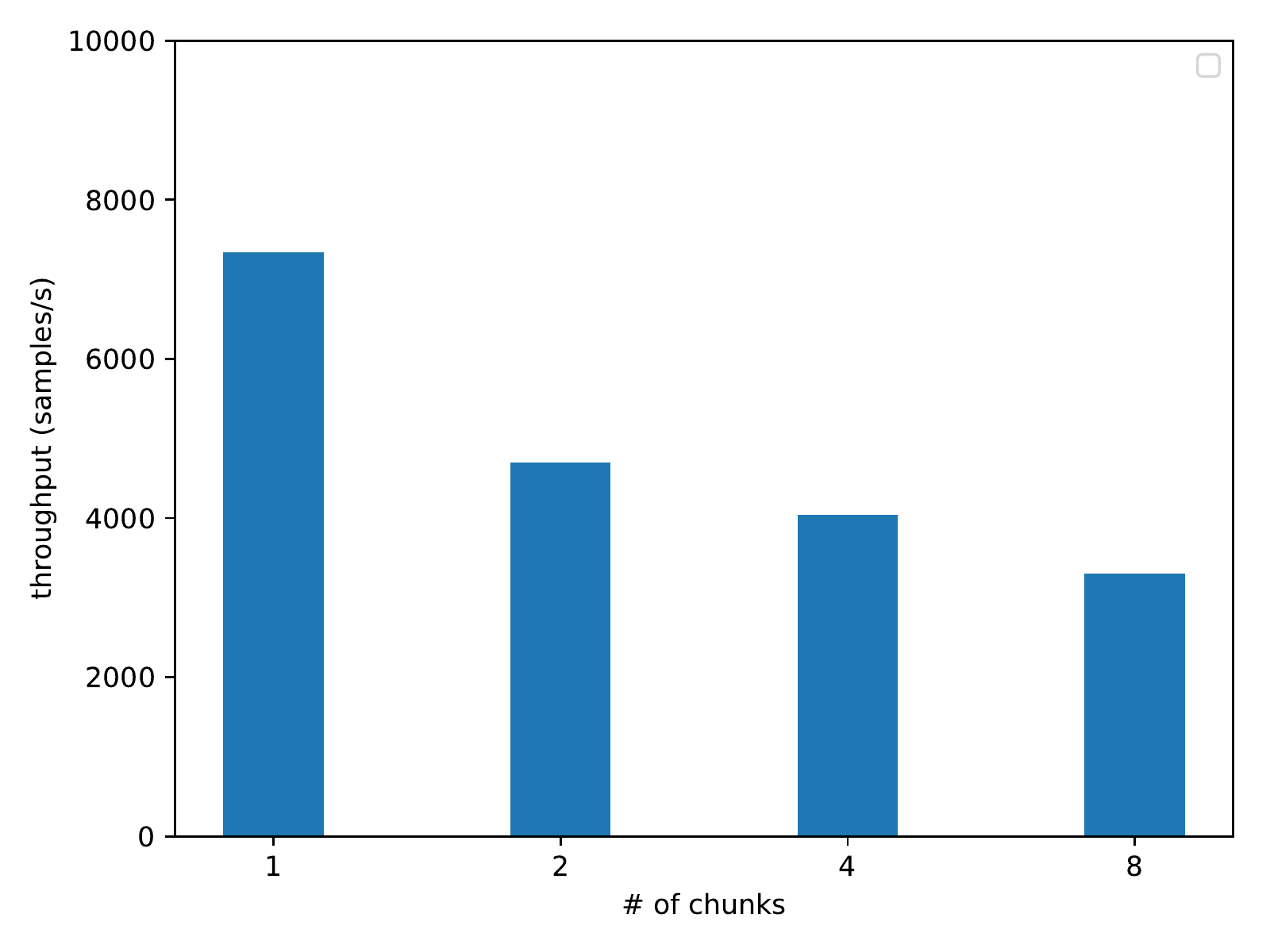}}
    \caption{The throughput varying with the number of chunks in pipelined overlapping}
    \label{fig:pipe_overlapping}
\end{figure}

From the result, we can observe that the inter-node-communication cost is roughly a few times larger than the sum of intra-node communication and expert forward propagation. Motivated by such a relationship in time cost, it is possible to overlap communication cost and computational cost.

In order to verify this idea, we utilize the pipeline mechanism to parallelize the execution of communication and computation on different hardware resources, i.e., GPU and NIC. We evaluated the throughput in a varying number of chunks. The results are shown in Figure \ref{fig:pipe_overlapping}. Unfortunately, no matter how we manipulate the chunk size, the performance still cannot improve. We argue that the performance degradation is due to the increase of more All2All operations. As we know from Section \ref{sec:breakdown} that the All2All operation is non-trivial. Although communication and communication are overlapped in some degree, the number of All2All operations are largely increased due to that the number of All2All communication operations inside the MoE layer increases linearly with respect to the number of chunks. This provides a new angle to understand the overhead of the All2All communication in the MoE layer.

\end{document}